\documentclass[conference]{IEEEtran}
\IEEEoverridecommandlockouts

\usepackage[utf8]{inputenc}
\usepackage{booktabs}
\usepackage{multirow}
\usepackage{graphicx}
\usepackage{threeparttable}
\usepackage{booktabs}
\usepackage{multicol}
\usepackage{float}
\usepackage{balance}
\usepackage{color}
\usepackage{algorithmic}
\usepackage{textcomp}
\usepackage{times}
\usepackage{gensymb}
\usepackage{booktabs}
\usepackage{array}
\usepackage[ruled,vlined]{algorithm2e}
\usepackage{tikz}
\usepackage[inline]{enumitem}
\usepackage{blindtext}
\usepackage{comment}
\usepackage[utf8]{inputenc}
\usepackage[T1]{fontenc}
\usepackage{xcolor}

\usepackage{amsmath, amssymb, amsfonts}
\usepackage{lipsum}
\usepackage{graphicx}
\usepackage{caption}
\usepackage{float}
\usepackage{amsmath}  
\usepackage{bm}       
\usepackage{cite}

\usepackage{algorithmic}

\def\BibTeX{{\rm B\kern-.05em{\sc i\kern-.025em b}\kern-.08em
    T\kern-.1667em\lower.7ex\hbox{E}\kern-.125emX}}
\begin{document}

\title{GuangMing-Explorer: A Four-Legged Robot Platform for Autonomous Exploration in General Environments\\
\thanks{* Corresponding Author}
\thanks{This work was supported by the Director Foundation of Guangdong Laboratory of Artificial Intelligence and Digital Economy(SZ)(25420001,24420004) and National Natural Science Foundation of China(42101445).}
}

\author{\IEEEauthorblockN{Kai Zhang}
\IEEEauthorblockA{\textit{1. Guangdong Laboratory of Artificial} \\ \textit{ Intelligence and Digital
Economy(SZ)} \\
\textit{2. Shenzhen University}\\
Shenzhen, China \\
zhangkai@gml.ac.cn}
\and
\IEEEauthorblockN{Shoubin Chen*}
\IEEEauthorblockA{\textit{1. Guangdong Laboratory of Artificial} \\ \textit{ Intelligence and Digital
Economy(SZ)} \\
\textit{2. Shenzhen University}\\
Shenzhen, China \\
chenshoubin@gml.ac.cn}
\and
\IEEEauthorblockN{Dong Li}
\IEEEauthorblockA{\textit{Guangdong Laboratory of Artificial} \\ \textit{ Intelligence and Digital
Economy(SZ)} \\
Shenzhen, China \\
doongli@ieee.org}
\and
\IEEEauthorblockN{Baiyang Zhang}
\IEEEauthorblockA{\textit{Guangdong Laboratory of Artificial} \\ \textit{ Intelligence and Digital
Economy(SZ)} \\
Shenzhen, China \\
zhangbaiyang@gml.ac.cn}
\and
\IEEEauthorblockN{Tao Huang}
\IEEEauthorblockA{\textit{Guangdong Laboratory of Artificial} \\ \textit{ Intelligence and Digital
Economy(SZ)} \\
Shenzhen, China \\
tao@gml.ac.cn}
\and
\IEEEauthorblockN{Zhouhong Cai}
\IEEEauthorblockA{\textit{Guangdong Laboratory of Artificial} \\ \textit{ Intelligence and Digital
Economy(SZ)} \\
Shenzhen, China \\
caizhouhong@gml.ac.cn}

\and
\IEEEauthorblockN{Zehao Wu}
\IEEEauthorblockA{\textit{1. Guangdong Laboratory of Artificial} \\ \textit{ Intelligence and Digital
Economy(SZ)} \\
\textit{2. Shenzhen University}\\
Shenzhen, China \\
wuzehao@gml.ac.cn}
\and
\IEEEauthorblockN{Jiasheng Chen}
\IEEEauthorblockA{\textit{1. Guangdong Laboratory of Artificial} \\ \textit{ Intelligence and Digital
Economy(SZ)} \\
\textit{2. Shenzhen University}\\
Shenzhen, China \\
chenjiasheng@gml.ac.cn}
\and
\IEEEauthorblockN{Bo Zhang}
\IEEEauthorblockA{\textit{Shenzhen University}\\
Shenzhen, China \\
zhangbo@gml.ac.cn}
}

\maketitle

\begin{abstract}

Autonomous exploration is a fundamental capability that tightly integrates perception, planning, control, and motion execution. It plays a critical role in a wide range of applications, including indoor target search, mapping of extreme environments, resource exploration, etc. Despite significant progress in individual components, a holistic and practical description of a completely autonomous exploration system, encompassing both hardware and software, remains scarce.
In this paper, we present GuangMing-Explorer, a fully integrated autonomous exploration platform designed for robust operation across diverse environments. We provide a comprehensive overview of the system architecture, including hardware design, software stack, algorithm deployment, and experimental configuration. Extensive real-world experiments demonstrate the platform's effectiveness and efficiency in executing autonomous exploration tasks, highlighting its potential for practical deployment in complex and unstructured environments.

\end{abstract}


\section{Introduction}
Autonomous exploration is a challenging yet crucial capability for robots operating in unknown environments. It allows the robot to gather essential information, such as traversability and occupancy maps, which form the basis for high-level tasks. The complexity of this task stems from the need to simultaneously perform two demanding processes in real time: incrementally building an accurate representation of the environment and generating feasible, collision-free paths to guide further exploration. Autonomous exploration has broad applications, including search and rescue~\cite{delmerico2019current}, surveillance~\cite{wang2020intelligent}, and planetary exploration~\cite{li2023special}, etc.


Moreover, the performance of an exploration system depends not only on the algorithms but also on the robustness and adaptability of the robotic platform itself~\cite{arm2019spacebok,moosavian2006design}. An effective platform must be capable of navigating and exploring diverse terrains, ranging from flat indoor surfaces to uneven outdoor environments~\cite{kolvenbach2022traversing}. In this paper, we present the design of our exploration platform, including sensor selection and system configuration, optimized for operation across a variety of complex and unstructured environments

To achieve accurate and efficient autonomous exploration, researchers have proposed a diverse set of methods, including classic sampling-based techniques~\cite{lavalle2001rapidly, karaman2011sampling}, information-gain-driven strategies~\cite{elfes1995robot}, frontier-based approaches~\cite{yamauchi1997frontier},
and more recent learning-based frameworks~\cite{cao2024deep}. The effectiveness of each method depends on several factors, such as the type of sensors used, the robot's motion model, environmental complexity, and the specific exploration strategy employed.
Each class of exploration algorithms presents distinct strengths and limitations. For instance, Rapidly-exploring Random Tree (RRT)-based methods~\cite{lavalle2001rapidly} explore the traversable space by growing a tree structure and selecting branches with the highest expected utility as the next viewpoints. These methods are computationally efficient and suitable for large-scale spaces due to their structural simplicity. However, they often miss smaller or narrow unexplored regions. On the other hand, methods that aim for complete coverage or fine-grained information gain may suffer from inefficiencies or even lead to suboptimal behavior in cluttered or constrained environments, such as getting trapped in local minima or repeatedly exploring low-value regions.



Many existing exploration methods determine the next best viewpoint by exhaustively comparing all candidate points, which becomes computationally prohibitive as the number of candidates grows. To address this, a series of hierarchical frameworks have been proposed~\cite{cao2021tare, long2024hphs,liang2024hdplanner}, which reduce computational complexity by narrowing the comparison scope. These frameworks decouple exploration into global and local planning modules, where the global planner provides directional guidance and the local planner focuses on fine-resolution mapping in the robot's immediate vicinity. This separation significantly accelerates candidate evaluation. However, such frameworks often suffer from degraded mapping accuracy in large-scale environments due to accumulated odometry drift.

In this work, we design an enhanced hierarchical exploration framework building upon TARE~\cite{cao2021tare} for our four-legged robot platform to achieve both high efficiency and mapping accuracy. We introduce a spatial-temporal calibration mechanism and integrate a more robust LiDAR-based odometry system to mitigate drift over extended trajectories. These improvements result in more accurate environmental representations and enable more reliable path planning, thereby enhancing both exploration efficiency and operational safety. Finally, we validate the proposed framework through a series of real-world experiments, demonstrating its practical effectiveness 

In summary, the key contributions include:
\begin{enumerate*}
\item We present a fully integrated autonomous exploration platform, featuring an efficient and accurate hierarchical framework capable of operating in diverse and unstructured environments.
\item We design a practical spatial-temporal calibration method and enhance the processing pipeline by incorporating robust LiDAR-based odometry, significantly improving mapping accuracy.
\item We validate the developed system through extensive experiments conducted in real-world environments, demonstrating its effectiveness and efficiency.
\end{enumerate*}


\section{Related Works}

\subsection{Autonomous Exploration Methods}

Since the influential work on frontier-based exploration~\cite{yamauchi1997frontier}, autonomous exploration strategies for 3D reconstruction have generally been classified into two main paradigms: frontier-based methods~\cite{yamauchi1997frontier} and sampling-based methods~\cite{zhu2021dsvp, dang2020graph}. Each paradigm offers distinct advantages and faces inherent limitations. Frontier-based approaches are well-suited for large-scale environments comprising multiple subregions, as they systematically guide the robot toward the boundaries between known and unknown areas. However, their frequent inter-region transitions often reduce efficiency due to suboptimal global path planning.
Conversely, sampling-based methods, such as those based on RRT, excel at detailed exploration within individual regions by effectively searching the configuration space. However, these methods tend to suffer from local entrapment in large-scale environments, resulting in incomplete coverage~\cite{selin2019efficient}.
To overcome these limitations, hybrid strategies have been proposed, leveraging the strengths of both paradigms. For example, some frameworks employ frontier-based approaches for global planning while using sampling-based methods for local viewpoint selection~\cite{selin2019efficient}. Others integrate sampling techniques into frontier-based strategies to reduce computational overhead and improve performance~\cite{cao2021tare}.
Recognizing that no single strategy consistently yields optimal performance across diverse scenarios, this paper proposes a unified framework that combines these complementary paradigms within a cohesive exploration system to enhance robustness and adaptability for exploration tasks.

\subsection{Autonomous Exploration Platform}
Autonomous exploration is a multifaceted task that involves the integration of hardware design, control systems, sensing technologies, local planning, and high-level exploration strategies. As such, the development of a universal and adaptable exploration platform is critical to support research and deployment across diverse environments. Compared to UAV-based exploration methods~\cite{zhou2021fuel, xu2021autonomous}, ground-based exploration presents more challenges due to factors such as terrain variability, traversability constraints, and physical limitations related to robot size and mobility.

While several prior works have proposed software frameworks for autonomous exploration~\cite{cao2022autonomous}, many of them focus on algorithmic components and provide limited insight into the practical deployment of these systems on real-world robotic platforms. In this paper, we aim to bridge this gap by presenting a comprehensive overview of our platform configuration, detailing both the software stack and the hardware setup. Our goal is to facilitate the development of fully functional autonomous ground explorers that can be readily adapted and deployed in complex, real-world environments.

\section{GuangMing-Explorer}

\begin{figure*}[htbp]
  \centering
  \includegraphics[width=0.8\linewidth]{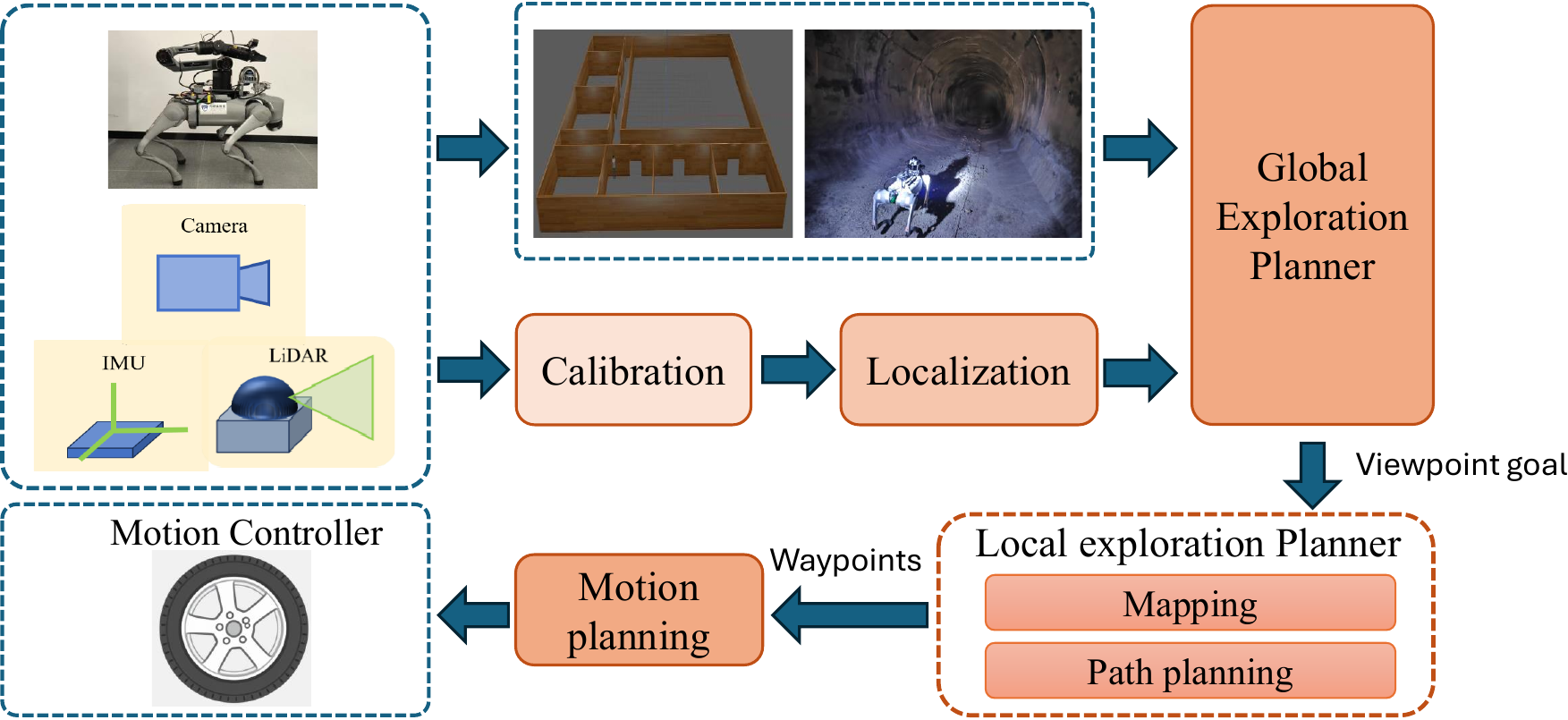}
  \caption{Pipeline for autonomous exploration on our four-legged robot platform.}
  \label{fig:overview}
\end{figure*}

An overview of the GuangMing-Explorer system is illustrated in Fig.~\ref{fig:overview}. The robot, equipped with multiple sensors is deployed to autonomously explore various environments, gather spatial data, and construct corresponding maps. The exploration system is composed of several interdependent modules: sensor calibration, robot localization, exploration (including both global and local planning), and motion control. These modules operate collaboratively to enable seamless and autonomous navigation and mapping.

In following sections, we begin by introducing the hardware (Section~\ref{sec:plat}) and software configuration of our exploration platform (Section~\ref{sec:algo}). Subsequently, we detail the design and implementation of each module, offering a comprehensive perspective on the integration of system components from both hardware and software aspects.

\subsection{Platform} \label{sec:plat}
\begin{figure}[h]
  \centering
  \includegraphics[width=0.95\linewidth]{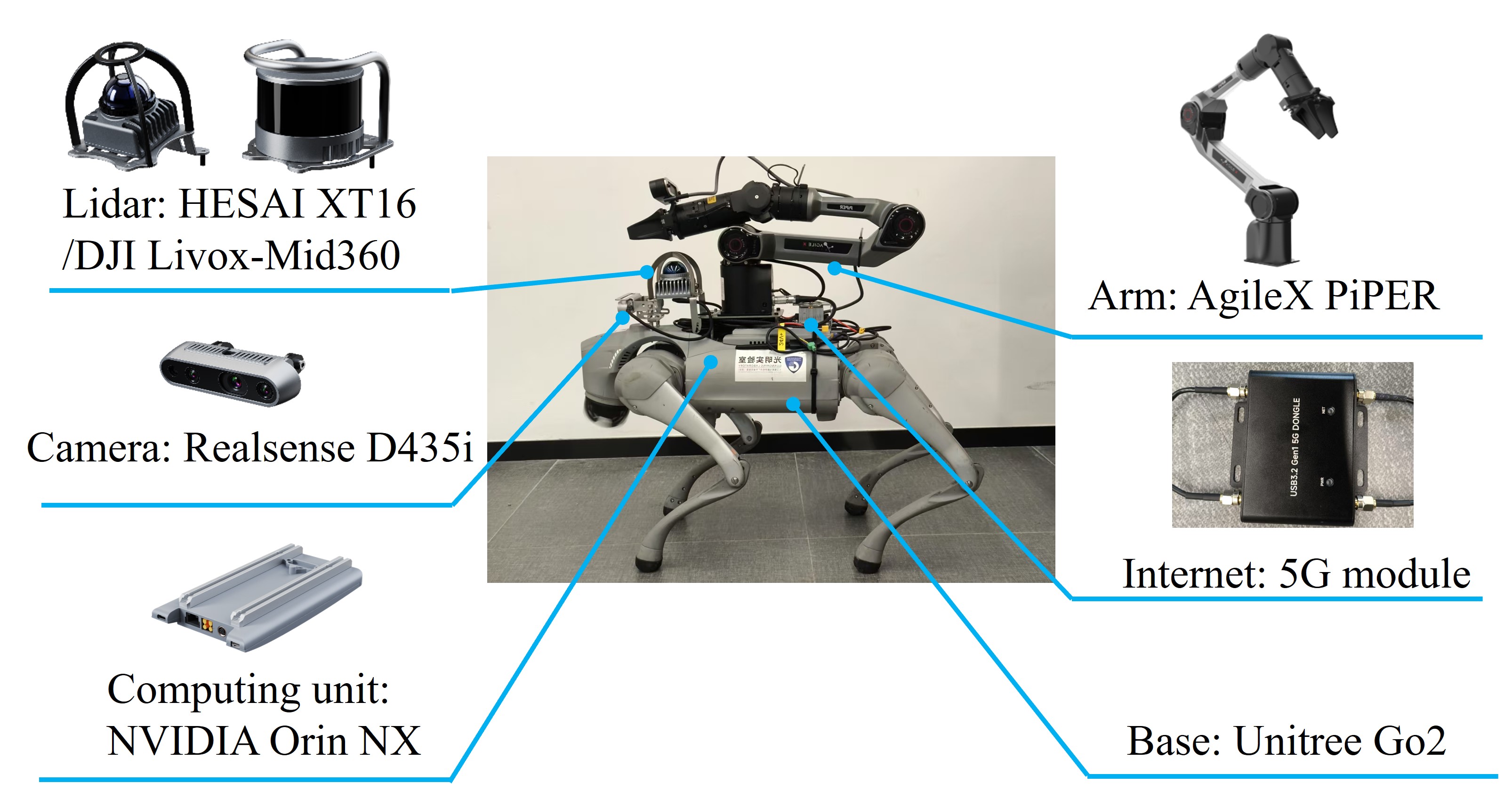}
  \caption{Configuration of the robotic platform.}
  \label{fig:platform}
\end{figure}

Our exploration platform is built upon the Unitree Go2 robot, as illustrated in Fig.~\ref{fig:platform}. The system is equipped with several sensors and modules designed to support robust autonomous exploration, including LiDAR units, a camera, a computing module, and a robotic arm. Specifically, the platform integrates two LiDAR sensors: the HESAI XT16, which provides long-range, sparse 3D perception, and the DJI Livox MID-360, optimized for dense, short-range measurements. For inertial sensing, we use the built-in IMU of the MID-360. To ensure reliable communication between the robot and remote terminals, a 5G communication module is installed, providing high-speed internet connectivity. The onboard computation is handled by an NVIDIA Jetson Orin NX (16 GB), which processes real-time data from LiDARs and an Intel Realsense D435i depth camera. All sensors (cameras and LiDARs) and controllers are connected to onboard computation units via 1 Gbps Ethernet cables to ensure high-bandwidth and low-latency communication. Additionally, the robot is equipped with a robotic arm, enabling basic physical interaction with the environment, such as opening doors or removing lightweight obstacles, further enhancing its autonomy in complex scenarios. This platform configuration allows the robot to navigate at speed of up to 3 m/s, supporting reliable operation across a wide range of environments, including indoor spaces, outdoor terrain, uneven surfaces, and underground caves, etc.

\subsection{Exploration algorithms}\label{sec:algo}

\subsubsection{Calibration} \label{sec:robot}
Accurate sensor calibration is essential for reliable autonomous exploration, particularly when fusing data from heterogeneous sensors. The calibration process involves two levels: individual sensor calibration and inter-sensor calibration to ensure precise temporal and spatial alignment.

For the Intel RealSense camera, intrinsic parameters such as focal length and principal point coordinates are typically provided by the manufacturer for image rectification. 

For LiDAR sensors, accurate time synchronization is critical to align point cloud data with other sensors' observation.
Two primary methods are employed for LiDAR time synchronization: Precision Time Protocol (PTP) and GPS-based synchronization. PTP operates by designating devices as either a master or a slave clock. It calculates clock offsets and network latency through timestamp exchanges, ensuring tight synchronization over a local network. In contrast, GPS synchronization relies on the Pulse Per Second (PPS) signal and the GPRMC time message. The PPS port emits a pulse every second, and the corresponding timestamp is transmitted through the data port in NMEA GPRMC format. Upon receiving the PPS pulse and decoding the GPRMC data, the LiDAR device aligns its internal clock to the GPS time and maintains synchronization through ongoing corrections. In our experiments, we apply the PTP calibration to synchronize the time of LiDARs.

Inter-sensor calibration is critical for accurate sensor fusion, as it determines the relative spatial transformation (rotation and translation) between different sensor coordinate frames. A common approach is to transform all sensor data into a unified coordinate system to enable consistent interpretation and fusion.
In our platform, we calibrate the LiDAR and camera by aligning their coordinate frames using a chessboard-based calibration method, following the approach described in~\cite{zhou2018automatic}. Specifically, a printed chessboard pattern is placed within the overlapping field of view of both the LiDAR and the camera. The camera captures an image of the scene, then the chessboard region is extracted. Simultaneously, the chessboard plane is isolated from the point cloud data.
By associating corresponding points observed by both sensors, a transformation matrix is computed via point set registration, representing the relative pose between the LiDAR and the camera. This transformation is subsequently used to align their measurements within a common reference frame. For setups involving more than two sensors, similar procedures are applied iteratively.
Once the calibration is complete, the platform is fully prepared for data collection.

\subsubsection{Localization}


Localization is a fundamental component of autonomous exploration, as it enables the robot to estimate its position and orientation in real time based on sensor observations. Accurate localization is essential for both mapping and motion planning.
In our system, we adopt Fast-LIO2~\cite{xu2022fast}, a high-performance LiDAR-inertial odometry framework specifically designed for fast and accurate state estimation in autonomous systems. Fast-LIO2 follows a map-centric approach, employing an Extended Kalman Filter to directly estimate the robot's state with respect to a global map. This strategy significantly mitigates the drift typically introduced by inertial measurements.
To further enhance computational efficiency, Fast-LIO2 utilizes an incremental KD-tree structure to dynamically update the local map, which accelerates point cloud registration and improves robustness in complex environments.
Through this localization module, the system generates a local occupancy map centered around the robot, indicating visible and occupied regions. This map serves as a foundation for both global and local planning modules.

\subsubsection{Exploration}

Effective autonomous exploration in large-scale, complex environments requires intelligent navigation strategies that optimize both coverage and efficiency. Our system builds upon the TARE~\cite{cao2021tare} exploration framework, which demonstrates robust performance across diverse environments but exhibits limitations when applied to legged-robot platforms due to their unique dynamic constraints. To make it practical in our platform, we present an enhanced version of TARE specifically adapted for quadruped robot systems.


Due to sensor range limitations and environmental occlusions, the robot's immediate perception is inherently constrained to a limited local region. To systematically expand this observable area, our approach employs a targeted viewpoint selection strategy inspired by the TARE framework. The local planner operates within a predefined square planning region encompassing both explored and unexplored areas.
The planning process begins by uniformly sampling candidate viewpoints along exploration frontiers at the boundary between known and unknown regions. These candidates are then evaluated based on their potential information gain, quantified by the observable area coverage. From this set, we select an optimal subset of high-value viewpoints as exploration waypoints, which enables efficient local area coverage.

While local path planning enables efficient exploration within a confined area, global planning is essential for navigating and covering large and complex environments. In our framework, global exploration is formulated as planning over a collection of discrete local regions, each represented as a cubic subspace.
The environment outside the current local map is first partitioned into evenly sized cuboid regions. Based on their exploration status, these regions are categorized into three types: unexplored, exploring, and explored. A region is marked as unexplored if it contains only unmapped surfaces, and as explored if it consists entirely of mapped areas. A region is labeled exploring if it contains a mixture of both explored and unexplored surfaces.
Global planning is executed by identifying the nearest "exploring" region and computing a path from the robot's current location to the centroid of that region. Upon completing local exploration within a region, the corresponding local map is integrated into the global map, and the robot transitions to the next adjacent unexplored or exploring region to continue its mission.
The exploration process concludes when no valid viewpoints remain, i.e., when all regions are classified as "explored," indicating complete coverage of the environment.


\subsubsection{Motion planning}

Once a local exploration path is generated, the robot must accurately follow it to reach the target viewpoint. For path tracking, we employ the Pure Pursuit algorithm~\cite{coulter1992implementation}, a widely used geometric control method known for its simplicity, efficiency, and robustness in mobile robotics and autonomous driving applications.
Pure Pursuit guides the robot by continuously steering it toward a look-ahead point on the planned path. This point is selected ahead of the robot's current position along the trajectory. The algorithm computes the required curvature by minimizing the angular deviation between the robot's heading and the direction to the look-ahead point. This approach enables smooth and responsive path following.
A critical parameter in Pure Pursuit is the look-ahead distance, which must be dynamically adjusted based on the robot's speed. A small look-ahead distance can cause unstable and oscillatory behavior, especially at high speeds, whereas a large distance may lead to significant tracking error and reduced path fidelity. In our setup, the look-ahead distance is proportional to the robot's speed, which can be observed from the sensors.
The motion controller receives the linear and angular velocity commands generated by the path tracker and executes them to control the robot's movement, ensuring safe and accurate navigation toward exploration goals.

\section{Experiment}

\subsection{Experiment Setup}


To evaluate the effectiveness of the proposed system, we conducted experiments in an office environment and an outdoor parking lot. An example is demonstrated in Fig.~\ref{fig:qual_res}.
The the office environment consists of multiple office rooms connected by a long corridor.
The outdoor parking environment is used primarily for qualitative validation due to the lack of detailed ground truth information. In both environments, the robot autonomously explored the space and generated an occupancy map based on the collected sensor data.

The algorithm runs on the onboard computing platform and follows similar experiments configuration as~\cite{cao2021tare}. 
The robot navigates with constrained velocities of 1.5 m/s maximum linear speed and 1.57 rad/s maximum angular speed to ensure stable movement.

For comprehensive performance assessment, we evaluate both exploration efficiency and accuracy. For the efficiency, we record the exploration completion time, average runtime and explored volume of the algorithm. Additionally, a high-precision 3D scan of the environment, obtained through manual operation, is used as the ground truth to compute the percentage of space explored. As for the accuracy, we calculate the cumulated localization error, the square root distance when the robot returns to its starting point.

\subsection{Overall Results}
\begin{table*}[h]
\centering
\captionsetup{width=0.9\linewidth}
\caption{\textsc{Quantitative results across five tests, reported as mean $\pm$ standard deviation}}
\label{tab:quan_res}
\begin{tabular}{|c|c|c|c|c|c|}
\hline
Test & Exploration   Time (s) & Runtime (s) & Explored Volume ($m^3$) & Traveling   Distance (m) & Localization   Error (m) \\ \hline
1    & 461.80                 & 0.60$\pm$0.05       & 1312.13(90.77\%)              & 236.56                   & 0.06                     \\ \hline
2    & 396.30                 & 0.60$\pm$0.07        & 1348.00(93.25\%)              & 239.98                   & 0.11                     \\ \hline
3    & 338.70                 & 0.53$\pm$0.06      & 1258.75(87.08\%)              & 219.60                   & 0.05                     \\ \hline
4    & 420.80                 & 0.59$\pm$0.06        & 1317.88(91.17\%)              & 244.72                   & 0.05                     \\ \hline
5    & 534.80                 & 0.61$\pm$0.05         & 1440.88(99.68\%)              & 258.17                   & 0.17                     \\ \hline
Overall & 430.48$\pm$65.65 & 0.59$\pm$0.03 & 1335.53$\pm$60.01(92.39\%) &239.81$\pm$12.49 & 0.09$\pm$0.05 \\ \hline
\end{tabular}

\end{table*}

\begin{figure}[h]
  \centering
  \includegraphics[width=\linewidth]{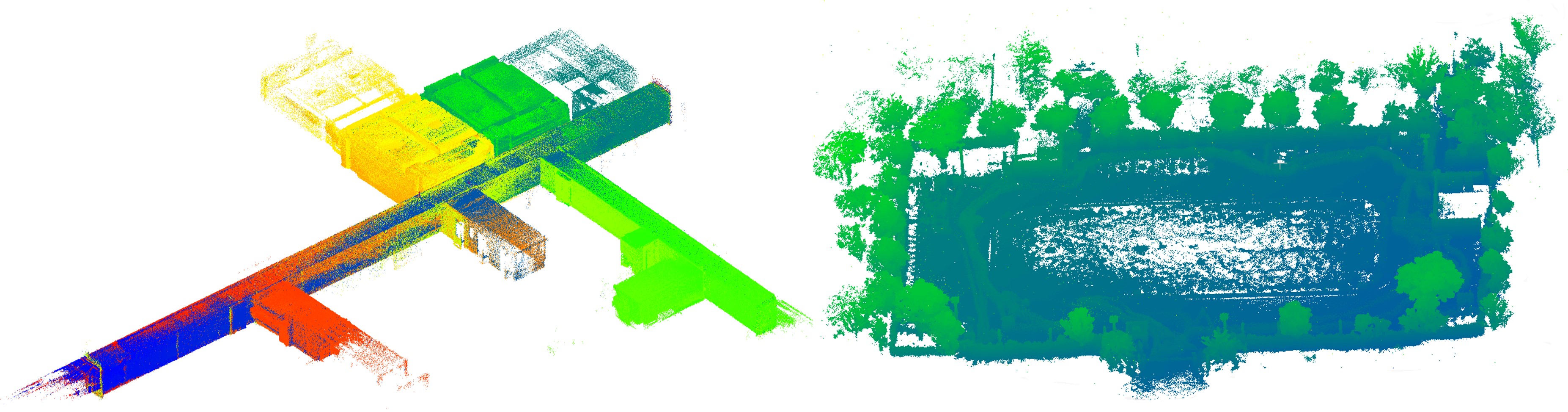}
  \caption{Point cloud of the explored environments: office~(left) and parking lot~(right).}
  \label{fig:qual_res}
\end{figure}

Five exploration trials were conducted in the office environment to evaluate the performance of the proposed autonomous exploration algorithm and robotic platform. The results are summarized in Table~\ref{tab:quan_res}, demonstrating the system's effectiveness.  The algorithm exhibits real-time performance, with an average path planning time of less than one second per iteration. The low standard deviation in execution time across trials confirms its computational stability. The constructed point cloud from both environments are shown in Fig.~\ref{fig:qual_res}.

Our comparison baseline, the original TARE algorithm, achieved 78.27\% coverage completeness while ours mapped 92.39\% of the environment on average. We found that due to the inaccurate mapping of original TARE, some entrance of rooms are considered blocked by mistakes so the robot ignored them during exploration. Additionally, we count the coverage variance between runs, where the small variation is attributed to the random sampling strategy used in selecting viewpoint candidates during planning.
Furthermore, the localization error remains small relative to the total travel distance, validating the accuracy of the modified LiDAR-based odometry. 

\subsection{Qualitative analysis on exploration}

\begin{figure}[h]
  \centering
  \includegraphics[width=0.8\linewidth]{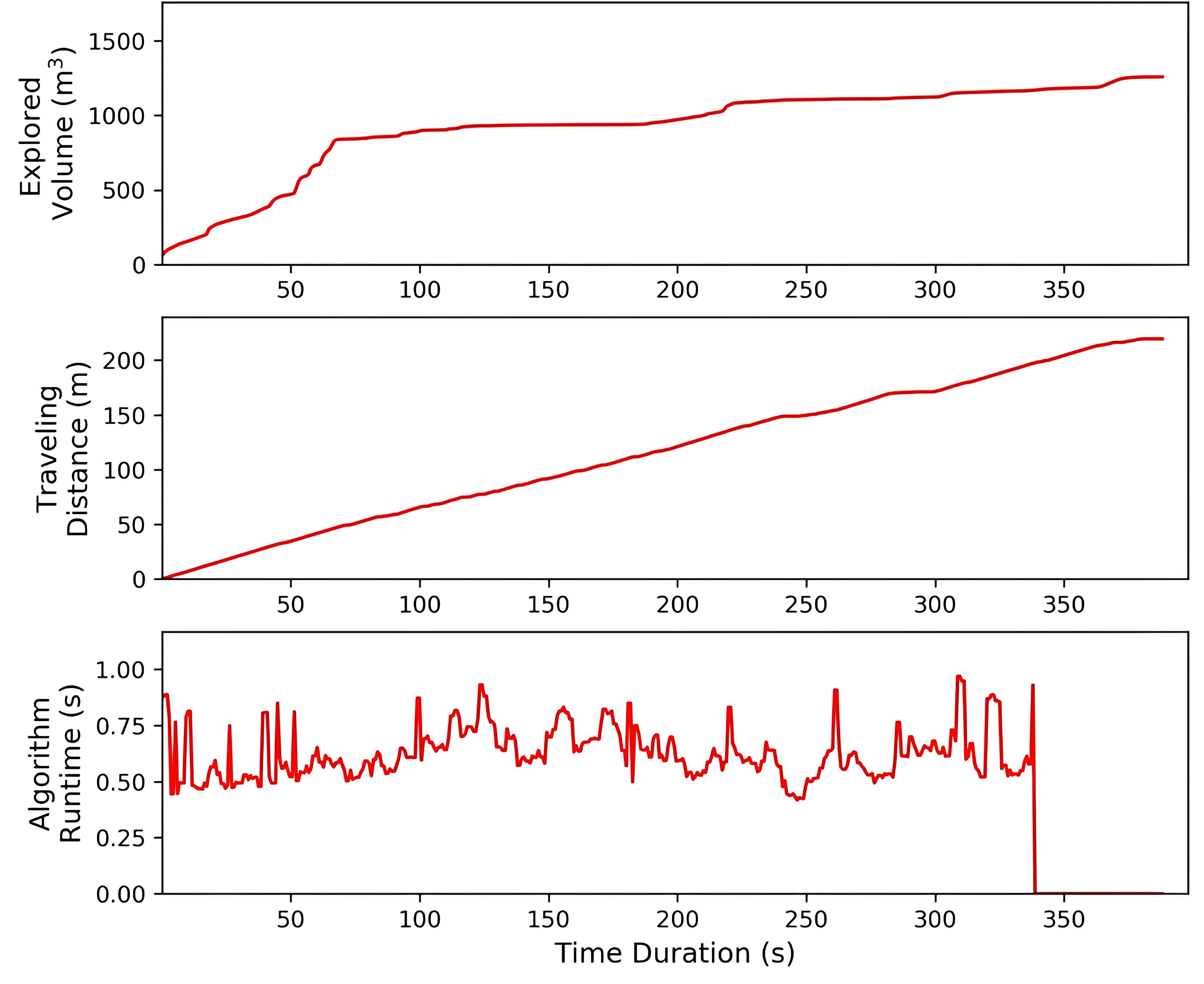}
  \caption{Temporal evolution of Test 3. The x-axis represents time while the y-axes (from top to bottom) shows the explored volume, traveled distance, and runtime per planning iteration.}
  \label{fig:test3-plot}
\end{figure}

To provide a detailed analysis of the experimental results, the explored volume, traveled distance, and algorithm runtime from Test 3 are plotted in Fig.~\ref{fig:test3-plot}. The explored volume curve shows that within the first 70 seconds (approximately 21\% of the total exploration time), the robot achieved approximately 67\% coverage, which highlights the efficiency of the proposed exploration algorithm. The traveled distance plot exhibits a steady increase, indicating continuous exploration without significant failures, demonstrating the effectiveness of the motion planner in generating feasible paths. The algorithm runtime plot reveals stable computational performance over time, despite minor fluctuations.

Nonetheless, several limitations can be observed. While the initial phase in explored volume curve demonstrates high efficiency, the extended time required to cover small unexplored regions indicates room for improvement. Specifically, the robot often revisits previously explored areas to access isolated unexplored pockets. Incorporating heuristic strategies to minimize such fragmented regions could reduce redundant movements and improve overall exploration efficiency.

\begin{figure}[h]
  \centering
  \includegraphics[width=0.8\linewidth]{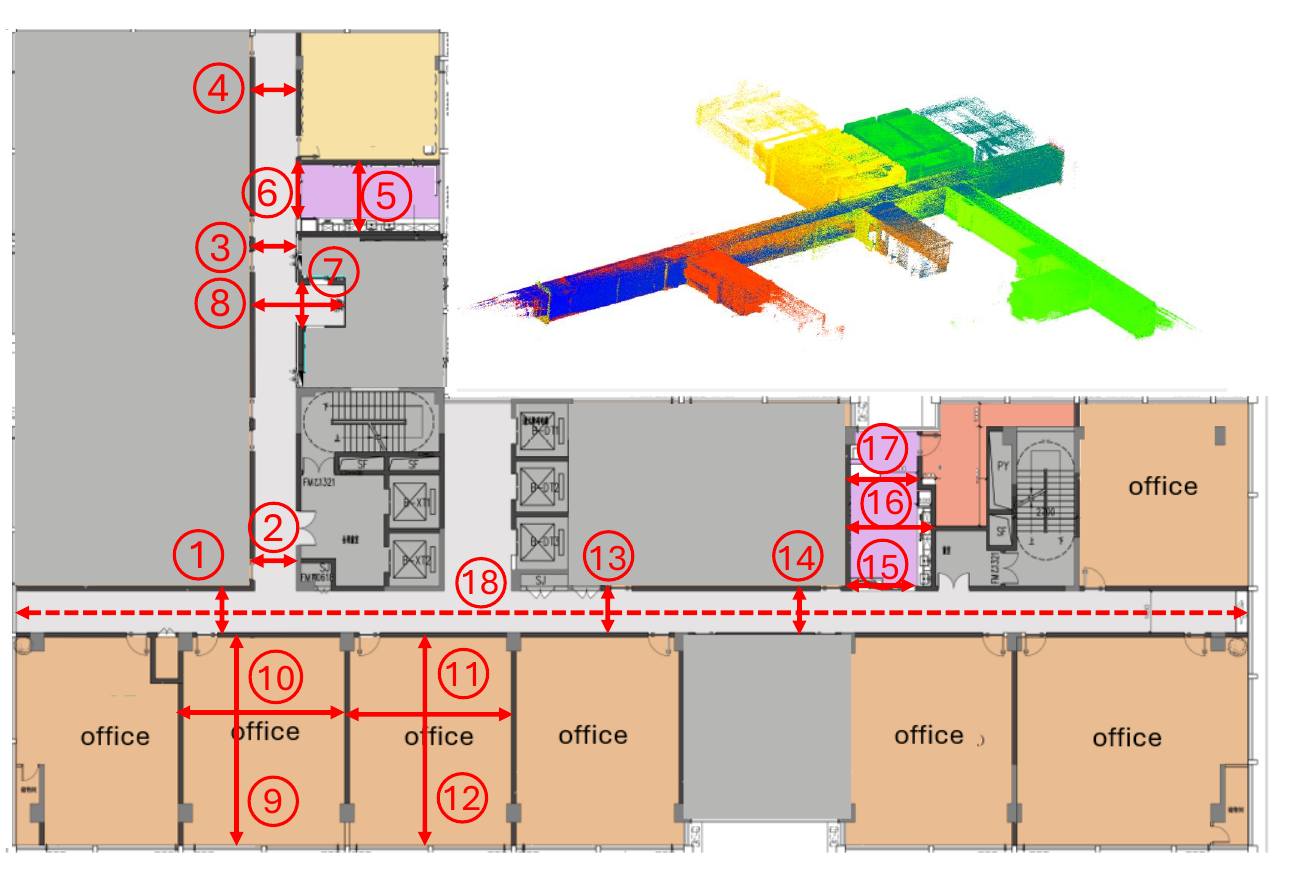}
  \caption{Evaluation of map accuracy. The red lines with numbers are selected lines for accuracy evaluation.}
  \label{fig:qual_map}
\end{figure}

\subsection{Qualitative analysis on mapping}

In the absence of a ground truth map, mapping accuracy was evaluated by comparing selected reference lines. As shown in Fig.~\ref{fig:qual_map}, 18 reference lines were manually selected from the layout map and precisely measured using a laser rangefinder. Corresponding line segments were then extracted from the generated 3D map for comparison. The average length of the selected lines was 637.8cm, with an average absolute error of 1.0cm (equivalent to 0.16\% relative error) and the maximum error is 2.5cm. These results demonstrate that the generated maps are sufficiently accurate for time-critical applications such as emergency response, where rapid deployment is prioritized over high mapping precision.

\section{Conclusion}

In this paper, we present GuangMing-Explorer, a robotic system comprising both hardware and software components, for autonomous exploration in diverse environments. We describe the sensor configuration and platform setup in detail, along with the key algorithms used for localization, exploration, and motion planning. The system is evaluated in office and parking environments, where experimental results demonstrate its stability and efficiency in completing exploration tasks. As future work, we plan to deploy the platform in more complex environments to further validate its performance and robustness. A reinforcement learning~\cite{miki2022learning} based locomotion controller is currently being prepared to improve adaptability and stability of the legged robot on those environments.


\bibliographystyle{IEEEtran}
\bibliography{IEEEfull}
\end{document}